\documentclass{article}

\PassOptionsToPackage{numbers, compress}{natbib}

\usepackage[preprint]{neurips_2024}

\usepackage[utf8]{inputenc}
\usepackage[T1]{fontenc}
\usepackage{hyperref}
\usepackage{url}
\usepackage{booktabs}
\usepackage{amsfonts}
\usepackage{nicefrac}
\usepackage{microtype}
\usepackage{xcolor}
\usepackage{algorithm}
\usepackage{algorithmic}
\usepackage{subcaption}
\usepackage{wrapfig}
\usepackage{ifsym}
\usepackage{diagbox}
\usepackage{float}
\usepackage{multirow}
\usepackage{epsfig}
\usepackage{marvosym}
\usepackage{setspace}
\usepackage{graphicx}
\usepackage{amsmath}
\usepackage{amssymb}
\usepackage{mathtools}
\usepackage{amsthm}
\usepackage{wasysym}
\usepackage{enumitem}
\usepackage{tcolorbox}
\usepackage{bbm}
\usepackage{makecell}
\usepackage{colortbl}
\usepackage{tikz}

\definecolor{baselinecolor}{gray}{.9}
\definecolor{myred}{rgb}{0.992,0.9576,0.932}
\definecolor{mydred}{rgb}{0.992,0.915,0.892}
\definecolor{mypink}{rgb}{1,0.95,0.962}
\definecolor{myyellow}{rgb}{0.99,1,0.78}
\definecolor{myredd}{rgb}{0.992,0.9076,0.63}
\definecolor{mydredd}{rgb}{0.96,0.72,0.72}
\definecolor{mypinkd}{rgb}{0.98,0.75,0.952}

\definecolor{myg4}{rgb}{0.98,0.98,0.98}
\definecolor{myg5}{rgb}{0.96,0.96,0.96}
\definecolor{myg6}{rgb}{0.94,0.94,0.94}
\definecolor{myg7}{rgb}{0.92,0.92,0.92}
\definecolor{myg8}{rgb}{0.9,0.9,0.9}
\definecolor{myg9}{rgb}{0.88,0.88,0.88}
\definecolor{myg10}{rgb}{0.86,0.86,0.86}
\definecolor{myg11}{rgb}{0.84,0.84,0.84}
\definecolor{myg12}{rgb}{0.82,0.82,0.82}
\usepackage{amssymb}
\usepackage{amsmath,amsfonts}
\usepackage{amsopn}
\usepackage{bm}
\usepackage{multirow}
\usepackage{flushend}
\usepackage{tabularx}
\newcommand{\Wings}{{\sc Wings}\xspace}
\newcommand{\Laws}{{\sc Laws}\xspace}

\newcommand{\ProbOpr}[1]{\mathbb{#1}}

\newcommand{\expect}[2]{%
\ifthenelse{\equal{#2}{}}{\ProbOpr{E}_{#1}}
{\ifthenelse{\equal{#1}{}}{\ProbOpr{E}\left[#2\right]}{\ProbOpr{E}_{#1}\left[#2\right]}}} 

\newcommand{\eat}[1]{}

\usepackage{xspace}
\newcommand{\ie}{{\it i.e.}\xspace}
\newcommand{\eg}{{\it e.g.}\xspace}

\title{\textsc{Wings}: Learning Multimodal LLMs \\without Text-only Forgetting}

\author{
  Yi-Kai Zhang$^{1,2,3}$\thanks{Work done during the internship at AI Business, Alibaba Group.} \ \ \ Shiyin Lu$^3$ \ \ \ Yang Li$^3$ \ \ \ Yanqing Ma$^3$ \ \ \ Qing-Guo Chen$^3$ \\\textbf{Zhao Xu}$^3$ \ \ \  \textbf{Weihua Luo}$^3$ \ \ \  \textbf{Kaifu Zhang}$^3$ \ \ \  \textbf{De-Chuan Zhan}$^{1,2}$ \ \ \  \textbf{Han-Jia Ye}$^{1,2}$\thanks{Corresponding author, email: yehj@lamda.nju.edu.cn.}\\
  $^1$School of Artificial Intelligence, Nanjing University\\
  $^2$National Key Laboratory for Novel Software Technology, Nanjing University\\ $^3$AI Business, Alibaba Group \\
}

\begin{document}

\maketitle

\begin{abstract}
Multimodal large language models (MLLMs), initiated with a trained LLM, first align images with text and then fine-tune on multimodal mixed inputs.
However, the MLLM catastrophically forgets the text-only instructions, which do not include images and can be addressed within the initial LLM.
In this paper, we present \Wings, a novel MLLM that excels in both text-only dialogues and multimodal comprehension.
Analyzing MLLM attention in multimodal instructions reveals that \textit{text-only forgetting} is related to the attention shifts from pre-image to post-image text. From that, we construct extra modules that act as the boosted learner to compensate for the attention shift.
The complementary visual and textual learners, like ``wings'' on either side, are connected in parallel within each layer's attention block.
Initially, image and text inputs are aligned with visual learners operating alongside the main attention, balancing focus on visual elements. Textual learners are later collaboratively integrated with attention-based routing to blend the outputs of the visual and textual learners.
We design the Low-Rank Residual Attention (LoRRA) to guarantee high efficiency for learners.
Our experimental results demonstrate that \Wings outperforms equally-scaled MLLMs in both text-only and visual question-answering tasks. On a newly constructed Interleaved Image-Text (IIT) benchmark, \Wings exhibits superior performance from text-only-rich to multimodal-rich question-answering tasks.
\end{abstract}

\section{Introduction}

Large Language Models (LLMs)~\cite{vicuna2023,du2022glm,jiang2023mistral,openai2020chatgpt,touvron2023llama,touvron2023llama2} are making significant strides toward Artificial General Intelligence (AGI) systems.
Multimodal Large Language Models (MLLMs), as a visual expansion of LLMs, have demonstrated astonishing performance in vision-related captioning~\cite{chen2023sharegpt4v,chen2015microsoft,li2022blip,li2023blip2}, understanding~\cite{bai2023qwenvl,chen2023internvl,dong2024ixc2,gpt4v,reid2024gemini1_5,step1v2023,wang2023cogvlm,zhu2023minigpt4}, and reasoning~\cite{alayrac2022flamingo,tsimpoukelli2021multimodal,wu2023visual,yang2023mmreact,zeng2022socratic}.
Common MLLMs build upon powerful pre-trained LLMs that take mixed textual and visual tokens as inputs. The visual ones are acquired using an image encoder and a projector.
We describe instructions processed by the LLM without images as \textit{text-only instructions}. In comparison, \textit{multimodal instructions} incorporate visual feature tokens into text-only sequences.
Modality fusing at the token level provides a flexible and effective pipeline for training MLLMs to comprehend visual information~\cite{lin2024moellava,liu2023llava,liu2024llavanext,xu2024llava_uhd}.
However, training on multimodal instructions seems to impair the pre-existing profound knowledge, especially making MLLM forget how to respond to text-only instructions like the initial LLM~\cite{lu2024deepseekvl,mckinzie2024mm1}.
MLLM experiences a drastic performance decline on text-only evaluation. We term it as the \textit{text-only forgetting} of MLLM.

\begin{figure}[t]
    \vspace{-20pt}
    \begin{center}
    \centerline{\includegraphics[width=1.1\textwidth]{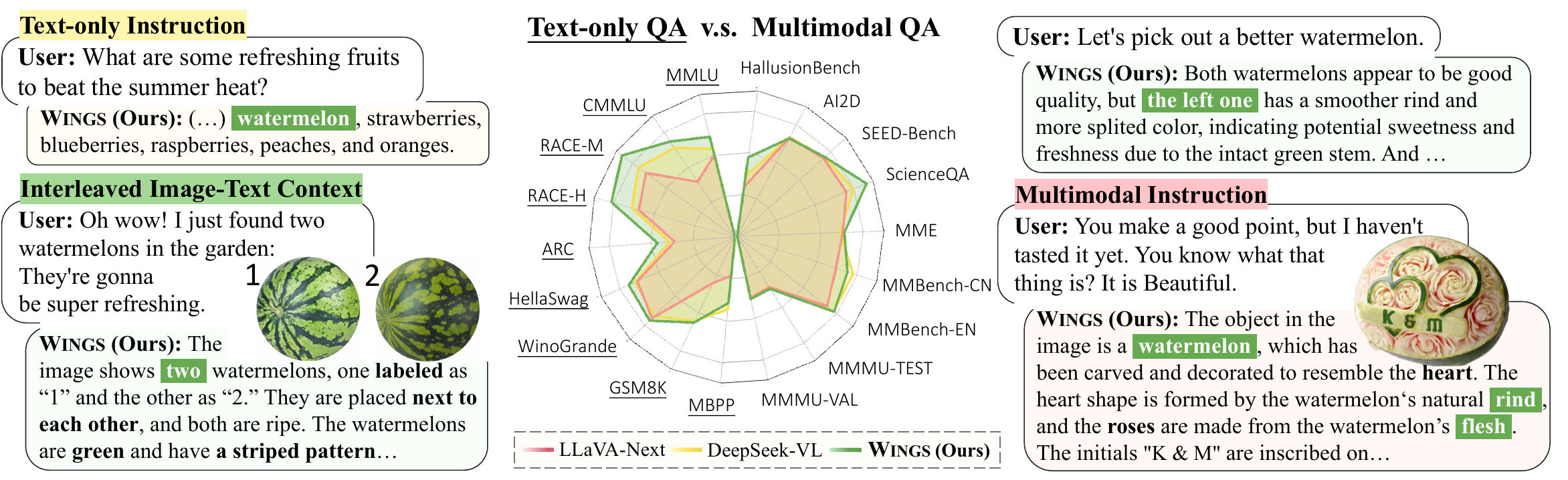}}
    \caption{\textbf{Examples of text-only and multimodal conversations.} From left to right: Interacting with MLLM through \textit{text-only and interleaved instructions}; Performance radar charts for \Wings, LLaVA-Next~\cite{liu2024llavanext}, and DeepSeek-VL~\cite{lu2024deepseekvl} in text-only and multimodal QA tasks, with dark green indicating \Wings with the comprehensive performance; Interacting with \textit{multimodal instructions}.}
    \label{fig:example}
    \end{center}
    \vspace{-35pt}
\end{figure}

In practical applications, MLLMs also require engaging in text-only or interleaved conversations.
As demonstrated in~\autoref{fig:example}, users often start with text-only inquiries and then, if not fully satisfied with the response, proceed to supplement questions with image content.
For multimodal instructions, MLLMs are still prompted to capture critical elements from text within a multimodal instruction, as images may provide redundant cues~\cite{mmstar,tomgpt,liu2024textmonkey}.
The first existing approaches replay extensive text-only or interleaved~\cite{laurencon2023idefics,DBLP:conf/nips/ZhuHAGDFYSW023} training data to mitigate catastrophic forgetting in MLLMs~\cite{chen2024allava,li2024miniGemini,lu2024deepseekvl,mckinzie2024mm1}.
However, increasing training data incurs additional computational overhead and data collection challenges.
Secondly, switching between LLM and MLLM based on whether images are included, as an intuitive solution, inevitably demands more deployment memory~\cite{DBLP:journals/corr/abs-2312-11514,DBLP:conf/sc/AminabadiRALLZRSZRH22} and is less cache-friendly in long vision-and-language interleaved conversations~\cite{he2024malmm,DBLP:journals/corr/abs-2401-02669,DBLP:conf/sc/RajbhandariRRSH21,Ren2023TimeChat}.
Therefore, it is crucial to train MLLM while preserving the text-only performance efficiently.

Given that the image feature tokens can be inserted at any position within the text sequence, we begin by examining the text before and after the inserted position.
Considering that MLLM's attention weights reflect the focus on tokens and influence the decision-making process, we first analyze the attention weights across each layer of the MLLM.
Specifically, for each layer, we compute the attention weight proportion on all text tokens before and after the inserted image, termed as \underline{L}ayer-level \underline{A}ttention \underline{W}eights (\Laws) of the before and after image text.
From this, we examine the dynamic of attention across all layers as MLLM-Laws.
Through training and sampling over $100$ diverse MLLMs, we find that a well-trained one with superior text-only performance shows a positive correlation of MLLM-\Laws between before and after image.
Given the similarity of feature space in the text surrounding the image, an MLLM's attention to both front and rear parts should be correspondingly similar.
A closer similarity indeed suggests a more minor disruption to the essential attention of MLLM.
Conversely, a negative correlation implies a shift in token attention across the image content, \ie, an MLLM overly concentrates on visual tokens and neglects textual ones.

Based on this observation, we propose \Wings, which introduces an extra module that acts as the boosted learner to compensate for the attention shift.
We integrate complementary visual and textual learners in parallel at each layer's attention block, with visual learners enhancing focus on visual tokens and textual learners on text, respectively.
In the first stage, visual features align with textual feature tokens, with all visual learners operating parallel to the main branch attention. The visual learners allocate some attention to visual tokens, mitigating the attention shift in the main branch.
Subsequently, textual learners are integrated in parallel. We implement soft routing based on shifted attention weights to harmonize the learning on visual and textual tokens.
We design the Low-Rank Residual Attention (LoRRA) as the architecture for learners to ensure high efficiency.
\autoref{fig:model} shows that the visual and textual learners on either side, like light feathers woven into ``wings''.
Experiments show that our \Wings comprehensively achieves superior performance in text-only under the same training condition and exceeds other equal-level MLLMs on multimodal benchmarks.
In addition, we construct the Interleaved Image-Text (IIT) benchmark with multi-turn evaluations towards a general mixed-modality scenario.
The samples are from text-only questions to strongly image-related conversations. \Wings achieve leading performance across various vision-relevance partitions.
Overall, our contributions are as follows:
(\textbf{1}) We claim and verify the text-only forgetting phenomenon of MLLM is related to the attention shift of cross-layer MLLM-\Laws before and after the image.
(\textbf{2}) \Wings construct the visual and textual learners and introduce a router based on shifted attention weights for collaborative learning to compensate for attention shifts.
(\textbf{3}) Experiments on text-only, visual-question-answering, and newly constructed Interleaved Image-Text (IIT) benchmarks demonstrate the comprehensive and versatile performance of \Wings.

\section{A Closer Look at Attention Shift in Multimodal LLMs}

In this section, we introduce the development from initialized LLM to MLLM. Next, we devise the MLLM-\Laws metric for representing attention shift and discuss the insights in building \Wings.

\subsection{Granting Sight to Large Language Models}

\textbf{Large Language Models (LLMs).} Even though existing Transformer-based~\cite{vaswani2017attention} models~\cite{DBLP:conf/slt/ChiCWHC0L21,liu2019roberta,raffel2020exploring,yang2019xlnet} like BERT~\cite{devlin2018bert} and OPT~\cite{zhang2022opt} have demonstrated profound language understanding capabilities, there has been a recent surge in powerful Generative Pre-trained Transformers (GPT)~\cite{brown2020language} under the auto-regressive language modeling paradigm.
Both public~\cite{jiang2023mistral,jiang2024mixtral,touvron2023llama,touvron2023llama2} and private~\cite{claude,openai2020chatgpt,openai2023gpt4,team2023gemini} solutions show remarkable progress in language comprehension and generation~\cite{DBLP:journals/corr/abs-2402-06196,wei2022emergent}.
These LLMs generally exceed a billion parameters, including pre-training~\cite{DBLP:journals/jmlr/ChowdheryNDBMRBCSGSSTMRBTSPRDHPBAI23,DBLP:journals/corr/abs-2301-00234,DBLP:conf/acl/0009C23,DBLP:journals/corr/abs-2001-08361}, supervised fine-tuning with instructions~\cite{DBLP:journals/corr/abs-2210-11416,DBLP:conf/iclr/SanhWRBSACSRDBX22,taori2023alpaca,wei2022finetuned}, and reinforcement learning from human feedback~\cite{DBLP:conf/nips/ChristianoLBMLA17,DBLP:conf/nips/Ouyang0JAWMZASR22,DBLP:conf/nips/StiennonO0ZLVRA20,DBLP:journals/corr/abs-1909-08593} on massive training data.

\begin{figure}[t]
    \vspace{-20pt}
    \begin{center}
    \centerline{\includegraphics[width=1.13\textwidth]{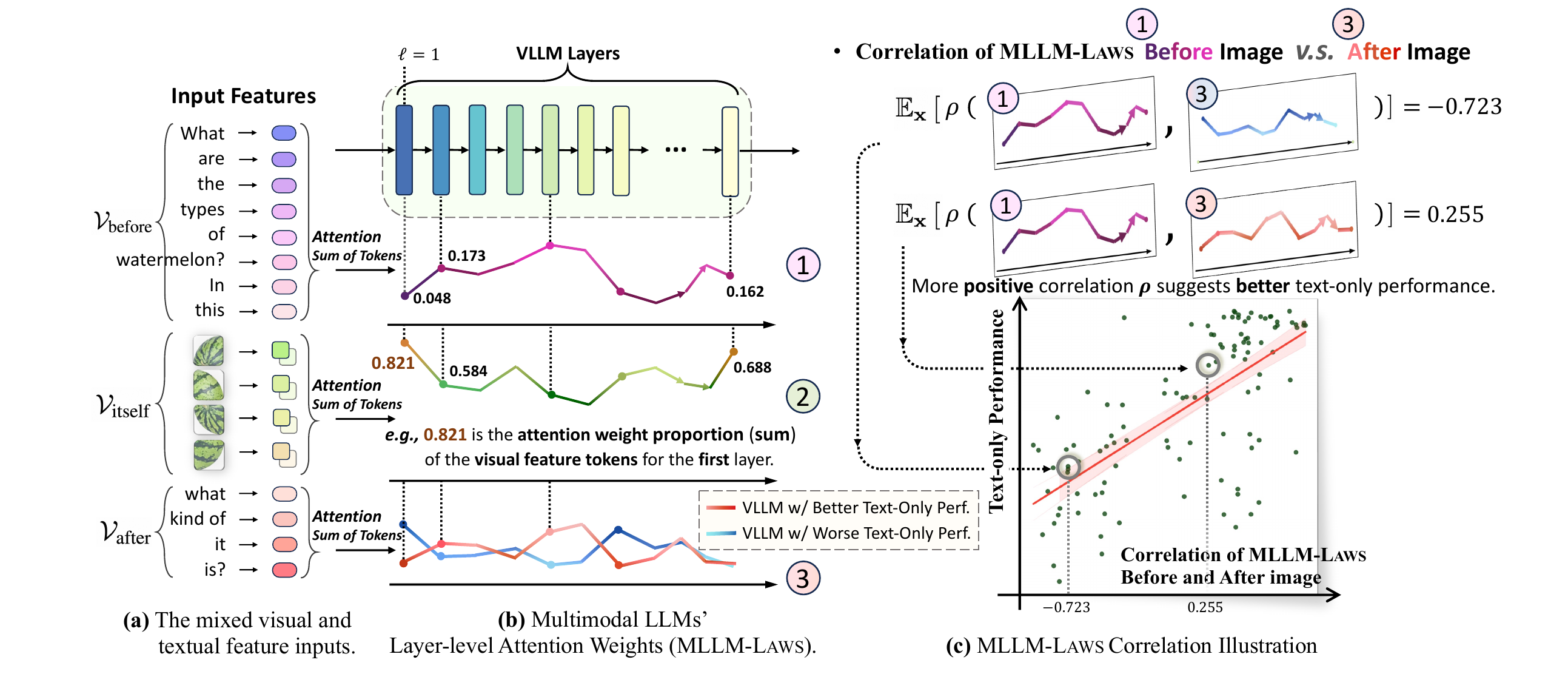}}
    \caption{\textbf{Illustration of mixed visual-and-textual inputs and the Layer-level Attention Weights ({\scshape Laws}) with its properties.} (a) The visual feature tokens from the visual encoder and projector are inserted into the textual feature sequence.
    (b) The attention weight proportion on textual tokens before-image, image-itself, and after-image across layers. The red curve is from the superior text-only MLLM, while the blue curve is from the inferior one.
    (c) Experiments on over $100$ MLLMs show a positive correlation from the $\boldsymbol{\rho}$ for MLLM-\Laws before and after the visual tokens ($x$-axis) to the text-only performance of the MLLM ($y$-axis).}
    \label{fig:illustration}
    \end{center}
    \vspace{-20pt}
\end{figure}

\textbf{Multimodal LLMs (MLLMs).}
Integrating visual inputs into foundational LLMs to create MLLMs is becoming increasingly popular~\cite{chen2023internvl,chen2024far,li2023otter,li2024miniGemini,ai2024yi}.
Unlike vision-centric multimodal frameworks~\cite{DBLP:journals/corr/abs-2206-02967,yuan2021florence} such as CLIP series~\cite{radford2021clip},
MLLMs aim to align new modality features as the input of LLM with an additional encoder~\cite{lin2023video,liu2023llava,liu2024llavanext,wu2023nextgpt,zhan2024anygpt,DBLP:journals/corr/abs-2310-01852}.
As illustrated in~\autoref{fig:illustration} (a), it enables the combined training of mixed multimodal tokens, facilitating rapid deployment across various applications~\cite{chu2023mobilevlm,chu2024mobilevlm,hong2023cogagent,liu2023llavaplus,wang2023cogvlm}.
One example of this pipeline is the LLaVA~\cite{liu2023llava} series, which integrates a CLIP vision encoder with a linear projection to Vicuna~\cite{vicuna2023} and innovatively introduces instruction-following training data.
Following this, some methods consider the richness of the vision-related training context~\cite{chen2023sharegpt4v,hu2023paperowl,li2023otterhd}, the scaled visual backbone~\cite{DBLP:journals/corr/abs-2312-14233,li2023monkey,lin2023sphinx}, or the enhanced connectors~\cite{cha2023honeybee,wei2023vary} to boost the visual effectiveness of MLLMs.
With commonplace text-only challenges in conversations, it is essential to enhance the language abilities of MLLMs~\cite{mckinzie2024mm1}.
The training process of MLLMs, as continued learning on newly introduced visual features, causes competitive modality shift~\cite{DBLP:conf/nips/GoelBBRVG22,DBLP:conf/nips/LiangZKYZ22,DBLP:conf/nips/0001XH23} and catastrophic text-only forgetting.
Recent studies acknowledge this issue, \eg, DeepSeek-VL~\cite{lu2024deepseekvl} suggests that supplementing additional text-only training data can mitigate this forgetting.
Others~\cite{DBLP:journals/corr/abs-2312-07533,mckinzie2024mm1} try to incorporate interleaved visual-textual data into training to retain language knowledge.
However, these methods are limited by training resources and data collection costs.
We aim to preserve or even boost performance with text-related training data as little as possible.
Some studies~\cite{jiang2024mixtral,li2024cumo,lin2024moellava,shazeer2017outrageously,DBRX,DBLP:journals/corr/abs-2403-13797,DBLP:conf/nips/ZhangHDZY23} also consider expanding the scalability of LLM, such as using Mixture-of-Expert (MoE) with numerous parallel FFNs in the Transformer block alongside a sparse gating network for efficient selection.
These methods, however, require a massive increase in training parameters.
In \Wings, the newly designed parallel learners of Low-Rank Residual Attention (LoRRA) are similar to MoE but with at least three orders of magnitude less in resource consumption.

\subsection{Capturing the Attention Shift with MLLM-{\scshape Laws}}
\label{sec:mllm_laws}

The significant decline in text-only performance is closely linked to the observed related shift during the training process.
Research on cross-modal learning~\cite{DBLP:conf/nips/DuanXZTZZ23,DBLP:conf/nips/LiSGJXH21,DBLP:conf/nips/LiangZKYZ22} shows that transferring to new modalities affects feature distribution, output values, and activation levels.
Considering attention weights highlight where MLLM's focus depends on visual or textual tokens for decision-making~\cite{film}, we investigate how attention shifts among \textit{different parts of the sequences}, mainly where divided by the visual feature tokens.
Specifically, we study over $100$ diverse MLLMs to uncover how attention is allocated to each part for a text-only better MLLM.
We take a closer look at the cross-layer dynamic curve of attention proportion on all text tokens {\em before} and {\em after} the inserted image.

For a instruction $\mathbf{x}$ and its hidden states in MLLM as $\mathbf{h} = \left[\mathbf{h}_1, \mathbf{h}_2, \cdots, \mathbf{h}_s\right]$ consisting of $s$ mixed visual and textual tokens.
Let $\mathrm{a}^{l}_{ij}$ represent the attention weight between the $i^{\text{th}}$ and $j^{\text{th}}$ tokens in the $l^{\text{th}}$ of the $L$-layers MLLM.
We have, for $\forall i$, $\sum_{j=0}^{s} \mathrm{a}^{l}_{ij} \left(\mathbf{h}^{l-1}\right) = 1$. 
As shown in~\autoref{fig:illustration} (a), since the sequence of flattened visual tokens is continuously interleaved with the textual sequence, we denote the index set of the visual tokens as $\mathcal{V}_{\text{itself}} = \left\{ v_{\text{start}}, v_{\text{start}} + 1, \cdots, v_{\text{end}} \right\}$. We refer to the textual sequence before the visual tokens as $\mathcal{V}_{\text{before}}$, and similarly, after the visual part as $\mathcal{V}_{\text{after}}$. For an MLLM with $L$ layers, we define the Layer-level Attention Weights (MLLM-\Laws) as:
\begin{equation}
    \text{\Laws}_{\,\mathcal{V}_{*}} = \left[ \mathrm{a}^{1}_{\mathcal{V}_{*}} , \mathrm{a}^{2}_{\mathcal{V}_{*}} \cdots, \mathrm{a}^{L}_{\mathcal{V}_{*}} \right]\;, \;\, \mathrm{a}^{l}_{\mathcal{V}_{*}} = \sum_{i=0}^{s} \sum_{j \in \mathcal{V}_{*}} \mathrm{a}^{l}_{ij} \left( \mathbf{h}^{l-1} \right)\;,
\end{equation}
where token index set $\mathcal{V}_{*}$ can be $\mathcal{V}_{\text{itself}}$, $\mathcal{V}_{\text{before}}$, or $\mathcal{V}_{\text{after}}$ as mentioned above, and for simplicity, we omit $\mathbf{h}^{l-1}$ in $\text{\Laws}_{\,\mathcal{V}_{*}}$.
In practice, $\text{\Laws}_{\,\mathcal{V}_{*}}$ characterizes the MLLM's attention on the current sequence $\mathcal{V}_{*}$ regarding the dynamic curve over all MLLM-layers.
As shown in~\autoref{fig:illustration} (b), the attention to the textual part initially increases and then decreases, while the trend for the visual one is often the opposite.
We find that when the MLLM forgets the text-only instructions, the \Laws of the textual sequence after the visual ones show a deviation from the initial trend of rising and then declining.
This implies a shift in the text following the image $\mathcal{V}_{\text{after}}$ compared to that preceding the image $\mathcal{V}_{\text{before}}$.
The dynamics labeled as \textcircled{3} in~\autoref{fig:illustration} (b) show the red curve for better text-only performance towards the worse blue one.
To quantify this, we compute the Pearson Correlation Coefficient~\cite{nguyen2020leep} between \Laws before and after the visual sequence. Formally,
\begin{equation*}
    \text{Attention Shift} = \mathbb{E}_{\mathbf{x}} \left[-{\rho}\left(\text{\Laws}_{\,\mathcal{V}_{\text{before}}}, \, \text{\Laws}_{\,\mathcal{V}_{\text{after}}}\right)\right] + 1\;.
\end{equation*}
Studying the attention shift of over $100$ diverse MLLMs, we find a positive correlation between the shift and the text-only performance degradation.
In~\autoref{fig:illustration} (c), each point represents a trained MLLM, and we demonstrate how the attention shift influences the text-only performance with the correlations.
Next, We focus on how to mitigate the shifted attention weights. Starting with \Laws we give the MLLM ``wings''.

\section{\textsc{Wings}: Flying to Generality with Low-Rank Residual Attention Learners}
From the attention shift, we seek a sufficiently reliable and convenient mechanism to address text-only forgetting.
The \Wings architecture is intuitive -- we construct visual and textual learners to mitigate shifted attention.
An attention-weight-based router dynamically adjusts the outputs of visual and textual learners to compensate for the main branch's attention outputs.
\Wings aims to excel in text-only and visual question-answering tasks with high generality.
In this section, we start with the typical training pipeline for MLLM (\autoref{sec:method_main_branch}).
Following this, we explore the motivation behind employing parallel modality learners and explain its implementation (\autoref{sec:method_learners}). Finally, we describe the training process for \Wings (\autoref{sec:method_training}).

\subsection{Revisit the Training Pipeline of the MLLM}
\label{sec:method_main_branch}

Following the mainstream architecture of MLLM, we take mixed visual and textual features as inputs.
For a one-turn conversation, the sequence of the visual feature tokens may appear at any position in the input $\mathbf{x}$. We represent the feature tokens as:
\begin{equation}
    \mathbf{x} = \left[ \mathbf{x}_{\texttt{V}}, \mathbf{x}_{\texttt{T}} \right] = \left[ \mathbf{h}_1, \cdots, \underbrace{\mathbf{h}_{v_{\text{start}}}, \mathbf{h}_{v_{\text{start}} + 1}, \cdots, \mathbf{h}_{v_{\text{end}}}}_{\text{visual features}}, \cdots, \mathbf{h}_{s} \right]\;,
\end{equation}
where we omit the superscript of layer-index $l$ for the $0^{\text{th}}$ layer. The $v_{\text{start}}$ and $v_{\text{end}}$ represent the starting and ending indices of the visual feature tokens, usually obtained through the vision encoder $\boldsymbol{\psi}$ and projector $\mathbf{W}_{\text{proj}}$, as $\mathbf{x}_{\texttt{V}} = \mathbf{W}_{\text{proj}} \cdot \boldsymbol{\psi} \left( \mathbf{x}_{\text{image}} \right)$.
Correspondingly, $\mathbf{x}_{\texttt{T}} = $
the remaining $0$ to $v_{\text{start}}$ and $v_{\text{end}}$ to length $s$ denote features of the textual instruction. We consider the posterior of the ground-truth answer as:
\begin{equation}
    \operatorname{Pr}\left( \mathbf{x}_{\texttt{a}} \mid \mathbf{x} \right) = \prod_{i=1}^{s} \mathbbm{1}_{\left[1, v_{\text{start}}\right) \cup \left(v_{\text{end}}, s\right]} \cdot \boldsymbol{\varphi} \left( \mathbf{h}_i \mid \left[ \mathbf{h}_1, \cdots, \mathbf{h}_{i-1} \right] \right)\;.
\end{equation}
Here, $\boldsymbol{\varphi}$ represents the main branch LLM, which consists of Transformer decoder layers~\cite{DBLP:conf/nips/VaswaniSPUJGKP17}.

\begin{figure}[t]
    \vspace{-10pt}
    \begin{center}
    \centerline{\includegraphics[width=0.72\textwidth]{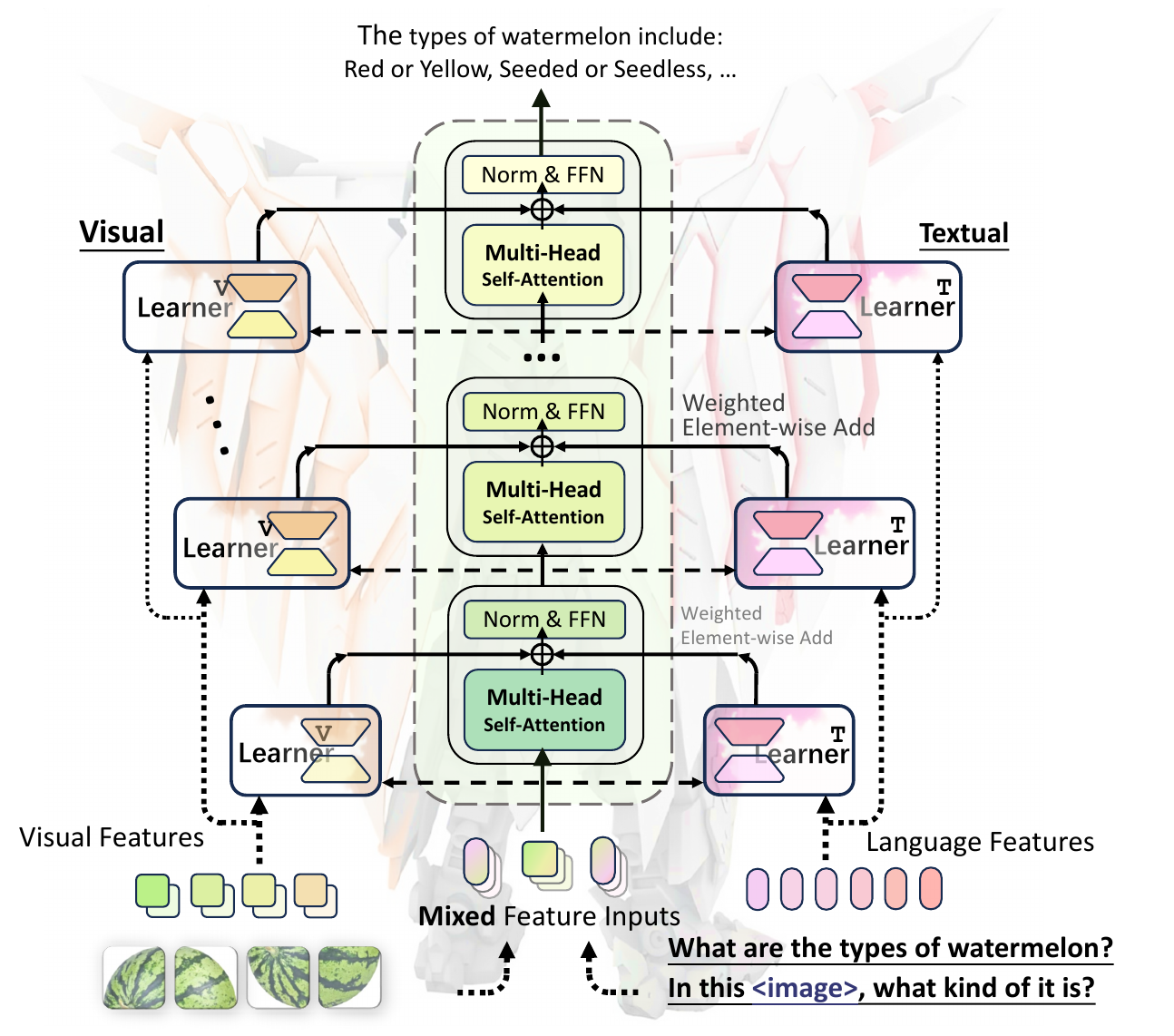}}
    \caption{\textbf{The {\scshape Wings} - model architecture.}
    We introduce extra modules parallel to the main attention, serving as boosted learners to compensate for the attention shift.
    We train the visual learners on one side, alleviating some shifted attention. Then, we collaboratively learn visual and textual learners based on routing shifted attention weights. They are like light feathers woven ``wings''.
    }
    \label{fig:model}
    \end{center}
    \vspace{-15pt}
\end{figure}

\subsection{Visual and Textual Learners Weave {\scshape Wings}}
\label{sec:method_learners}

\textbf{Motivation: Learning to mitigate the attention shift with modality-specific auxiliary structures.} As mentioned in~\autoref{sec:mllm_laws}, MLLM-Laws demonstrates the attention shift in the sequence following the visual features.
The shift results from excessive dependency on visual features.
This issue may stem from the insufficient alignment within mixed inputs~\cite{bai2023qwenvl,mmstar}, where new modalities can obscure existing knowledge.
It suggests adding a small, adjustable factor to the shifted mixed modality features and regulating unnecessary fluctuations in MLLM-\Laws.
Consequently, we aim to adopt an efficient, learnable module as the visual ``wing''.
Compared to the image-text mixed feature inputs of the main branch, it should focus on extracting visual information to share the burden of overly shifted attention.
The interaction between the current hidden state and visual features is conducted within this module.
Similarly, to balance the auxiliary function of the visual learner, we also construct a symmetrical textual learner.
Moreover, we should appropriately distribute the two learners across both modalities to operate collectively.

\begin{figure}[t]
    \vspace{-20pt}
    \begin{center}
    \centerline{\includegraphics[width=1.05\textwidth]{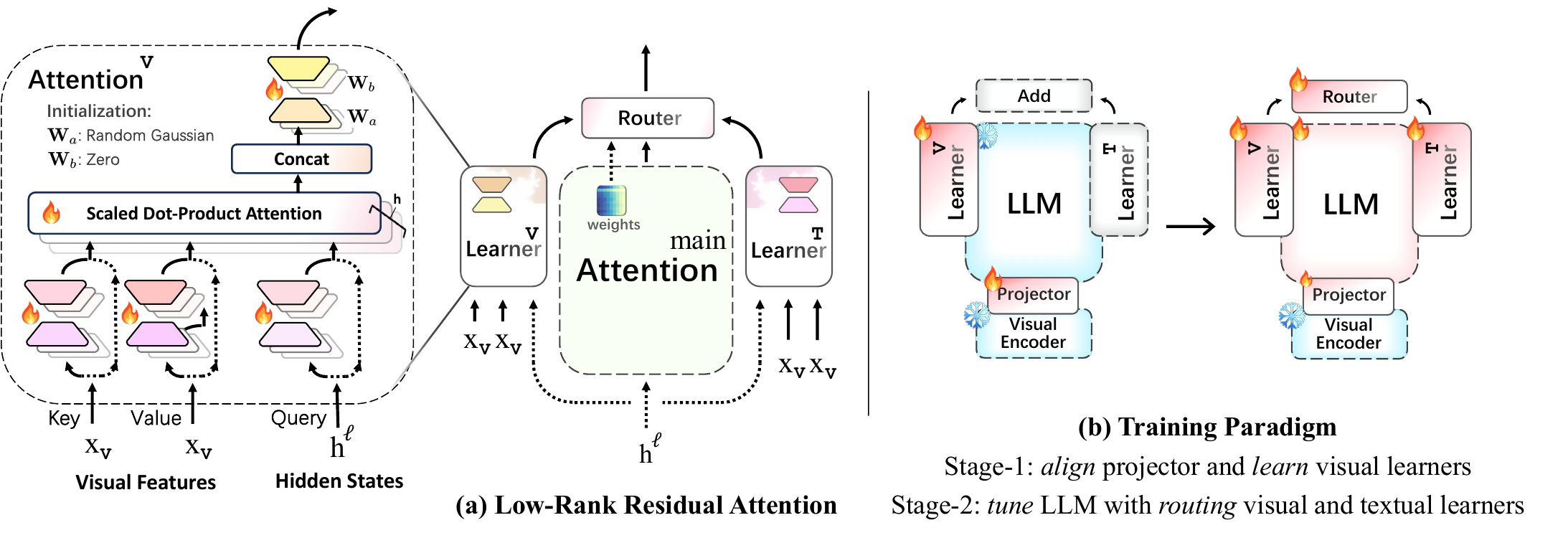}}
    \caption{\textbf{Illustrations of the detailed {\scshape Wings} structure, and training strategies.}
    \Wings is constructed by the Low-Rank Residual Attention (LoRRA) module where the previous hidden state acts as the \texttt{query} and the visual/textual features serve as the \texttt{key} and \texttt{value}. Training starts with visual learners and projectors, followed by the dynamic attention-based routing.}
    \label{fig:training}
    \end{center}
    \vspace{-20pt}
\end{figure}
\textbf{Structure: parallel learner \& router of attention outputs.}
To capture key information in shifted modalities while ensuring efficiency, we design a multihead Low-Rank Residual Attention (LoRRA) learner at every layer. It takes input from the hidden state and interacts with the initial visual or text-only feature.
The learner facilitates cross-cascading with the initial projected information. Specifically, for the $l^{\text{th}}$ layer, the visual/text-only learner is formulated as:
\begin{equation}
\small
\begin{aligned}
\mathsf{Learner}^{*} \left( \mathsf{Q} \text{=} \mathbf{h}^{l}, \mathsf{K}\text{,}\mathsf{V} \text{=} \mathbf{x}_{*} \right)_{* \in \left\{ \texttt{V}, \texttt{T}\right\}} = \operatorname{Softmax}\left(\frac{ \mathbf{h}^{l} \left( \boldsymbol{\mathbf{1}} + \mathbf{W}^{\mathsf{Q}} \right) \cdot \left( \mathbf{x}_{*} \left( \boldsymbol{\mathbf{1}} + \mathbf{W}^{\mathsf{K}}\right)\right)^{\top}}{\sqrt{d_{\text{head}}}}\right) \mathbf{x}_{*} \left( \boldsymbol{\mathbf{1}} + \mathbf{W}^{\mathsf{V}} \right) \mathbf{W}^{\mathsf{O}}\;,
\end{aligned}
\end{equation}
where the matrix $\mathbf{W}^{\mathsf{Q}}$, $\mathbf{W}^{\mathsf{K}}$, $\mathbf{W}^{\mathsf{V}}$, and $\mathbf{W}^{\mathsf{O}}$ is low-rank and is obtained by the dot product of $\mathbf{W}_a \in \mathbb{R}^{d \times \underline{d}}$ and $\mathbf{W}_b \in \mathbb{R}^{\underline{d} \times d}$, and $\underline{d}$ is relatively small enough.
$\boldsymbol{\mathbf{1}}$ is represented as the identity matrix.
Following LoRA~\cite{DBLP:conf/iclr/HuSWALWWC22}, LoRRA learners also employ random Gaussian initialization for $\mathbf{W}_a$ and sets $\mathbf{W}_b$ to zero. Given that $\mathbf{W}^{\mathsf{Q}}$ lacks a residual, the learner's output is zero at the beginning of training.
Multihead LoRRA preserves the effectiveness of the cross-attention structure and employs efficient low-rank mapping to reduce computational demands.
As shown in~\autoref{fig:model}, the visual and textual features are fed into their respective side learners, like two ``wings'' woven together.
The outputs of two learners from each layer are then weighted $\operatorname{sum}$ to the attention of the main branch.
As illustrated in the left part of~\autoref{fig:training}, a router receives attention weights to generate the balance weights of the visual and textual learners. In summary, we formulate the \Wings block as:
\begin{equation}
    \mathsf{Att}^{\textsc{Wings}} = \mathsf{Att}^{\text{main}} + \sum_{* \in \left\{\texttt{V}, \texttt{T} \right\}} 
    \mathsf{Router} \left( \mathbf{a} \right) \cdot \mathsf{Learner}^{*} \left(\mathbf{h}^{l}, \mathbf{x}_{*} \right)\;,
\end{equation}
where $\mathbf{a} \in \mathbb{R}^{s \times s}$ represents the attention weights of the current main branch.
The router receives the attention weights $\mathbf{a}$ and then processes through a single-layer $\mathsf{MLP}$ with $\operatorname{Softmax}$.

\subsection{Stable Training Recipe}
\label{sec:method_training}
The architecture of \Wings comprises four elements: vision encoder, projector, initialized LLM, and the learners with a router.
During the training process, the vision encoder is consistently fixed.
Firstly, we only fine-tune the projector and visual learners.
We primarily employ image-text pairs for visual alignment, while the outputs of visual learners are directly added to the main branch.
Subsequently, the LLM branch is updated with small steps.
Concurrently, textual learners are paralleled with visual learners on the attention block of LLMs.
The router learns to allocate visual and textual learners from the attention weights of the main branch.
At this stage, the textual and visual learners work better together to direct attention to the key tokens.
To summarize, \Wings prioritizes enhancing visual learners first. Subsequently, it ``spreads its wings'' by concurrently learning and routing visual and textual learners based on shifted attention weights.

\section{Experiments}

\label{sec:experimental_setup}
In this section, we first introduce the benchmarks for evaluating \Wings, including \autoref{tab:bench_text_only}: text-only forgetting on the same multimodal training data, \autoref{tab:bench}: comparison with general MLLMs, and \autoref{fig:ablation}: analysis on the Interleaved Image-Text (IIT) benchmark with varying levels of vision-related conversation. Following that, we outline the training details and configurations of the \Wings, and delve into experimental analysis across each benchmark. Moreover, we perform an ablation study on various learning rates with different training parts.

\textbf{Evaluation Setups.} We aim to assess through MLLM how much visual information is required for evaluation.
For example, generic multimodal instructions require MLLMs to strongly capture image aspects, whereas text-only instructions focus on the text. We introduce three types of benchmarks:

\begin{itemize}[itemsep=0pt,topsep=0pt,parsep=0pt,leftmargin=10pt]
    \item \textbf{Standard text-only benchmarks.} We are particularly interested in the text-only performance improvement of \Wings under the same training data and resource conditions. Different datasets including \textit{interdisciplinary exams} like MMLU~\cite{mmlu}, CMMLU~\cite{cmmlu}, ARC-Easy, ARC-Challenge~\cite{arc}, language \textit{understanding} and \textit{knowledge} such as WinoGrande~\cite{sakaguchi2019winogrande}, OpenbookQA~\cite{DBLP:conf/acl/BanerjeePMB19}, Race-Middle, Race-High~\cite{race}, WSC~\cite{clue}, CHID~\cite{chid}, \textit{reasoning} such as HellaSwag~\cite{hellaswag}, SIQA~\cite{siqa}, PIQA~\cite{piqa}, OCNLI~\cite{ocnli}, and \textit{math} and \textit{code}-related tasks such as GSM8K~\cite{gsm8k} and MBPP~\cite{mbpp} are comprehensively evaluated.
    \item \textbf{General multimodal benchmarks.} We evaluate on MMMU~\cite{yue2023mmmu}, MME~\cite{fu2023mme}, MMBench~\cite{liu2023mmbench} (MMB) in English (EN) and Chinese (CN), ScienceQA~\cite{lu2022scienceqa} for test (SciQA), SEED-Bench~\cite{li2023seed} for image part (SEED), AI2D~\cite{kembhavi2016ai2d} for test, and HallusionBench~\cite{guan2023hallusionbench} (HallB).
    \item \textbf{Our Interleaved Image-Text (IIT) benchmark} with diverse text-only, interleaved, and image-ralated multi-turn conversations.
    It includes sampling for MMLU, CMMLU, OpenbookQA, HellaSwag, MMMU, MMBench, SEED-Bench, and AI2D datasets.
\end{itemize}

\begin{table*}[t!]
\scriptsize
\vspace{-10pt}
\centering
\renewcommand{\arraystretch}{1.3}
\begin{tabular}{p{3.5em} p{5.2em} p{2em}<{\centering} p{2em}<{\centering} p{2em}<{\centering} p{2.8em}<{\centering} | >{\columncolor{mypink}}p{2em}<{\centering}<{\centering} p{2em}<{\centering} p{2em}<{\centering} >{\columncolor{myred}}p{2em}<{\centering}<{\centering}| >{\columncolor{mydred}}p{2em}<{\centering}<{\centering} p{2.8em}<{\centering} p{2.6em}<{\centering}}
\toprule
\multicolumn{2}{c}{\multirow{2}{*}{\diagbox[height=27pt,innerrightsep=10pt]{\textbf{Dataset}}{\textbf{Model}}}} & \raisebox{-1ex}{\tiny{\textbf{Vicuna}}} & \raisebox{-0.8ex}{\tiny{\textbf{Vicuna}}} & $\overset{\text{\tiny{}}}{\text{\raisebox{-0.8ex}{\tiny{\textbf{LoRA}}}}}_{\text{{\textbf{{Vicu.}}}}}$ & \raisebox{-0.8ex}{\tiny{\textbf{Vicuna}}}  & \raisebox{-1ex}{\tiny{\textbf{Qwen}}} & \raisebox{-0.8ex}{\tiny{\textbf{Qwen}}} &  $\overset{\text{\tiny{}}}{\text{\raisebox{-0.8ex}{\tiny{\textbf{LoRA}}}}}_{\text{{\textbf{{Qw.}}}}}$  & \raisebox{-0.8ex}{\tiny{\textbf{Qwen}}} & \raisebox{-1.3ex}{{\tiny{\textbf{{\scshape Wings}}}}} & \raisebox{-1ex}{\tiny{\textbf{Text-only}}} & \raisebox{-1ex}{\tiny{\textbf{Our}}} \\
 & & \raisebox{1.5ex}{\tiny{\textbf{LLM}}} & $\overset{\raisebox{0.5ex}{+}}{\text{\tiny{\textbf{CLIP}}}}$ & $\overset{\raisebox{0.5ex}{+}}{\text{\tiny{\textbf{CLIP}}}}$ & $\overset{\raisebox{0.5ex}{+}}{\text{\tiny{\textbf{SigLIP}}}}$ & \raisebox{1.5ex}{\tiny{\textbf{LLM}}} & $\overset{\raisebox{0.5ex}{+}}{\text{\tiny{\textbf{CLIP}}}}$ & $\overset{\raisebox{0.5ex}{+}}{\text{\tiny{\textbf{CLIP}}}}$ & $\overset{\raisebox{0.5ex}{+}}{\text{\tiny{\textbf{SigLIP}}}}$ & \raisebox{1.6ex}{\tiny{\textbf{(Ours)}}} & $\overset{\raisebox{0.5ex}{\text{\tiny{\textbf{Forgetting}}}}}{\text{\mbox{(\colorbox{mypinkd}{\parbox[c][0.4pt]{0.4pt}{\hspace{0.3pt}}} - \colorbox{myredd}{\parbox[c][0.4pt]{0.4pt}{\hspace{0.3pt}}})}}}$ & $\overset{\raisebox{0.5ex}{\text{\tiny{\textbf{Impro.}}}}}{\text{\mbox{(\colorbox{mydredd}{\parbox[c][0.4pt]{0.4pt}{\hspace{0.3pt}}} - \colorbox{myredd}{\parbox[c][0.4pt]{0.4pt}{\hspace{0.3pt}}})}}}$ \\
\midrule
\multirow{4}{*}{\textbf{Exam}} & MMLU & 51.18 & 51.12 & 48.89 & 50.63 & \textbf{60.86} & 50.83 & 59.67 & 51.16 & \underline{60.53} & \cellcolor{myg11} 9.70 & \cellcolor{myg10} 9.37 \\
& CMMLU & 38.60 & 38.29 & 37.24 & 38.73 & \underline{69.37} & 62.58 & 67.87 & 60.46 & \textbf{69.82} & \cellcolor{myg10} 8.91 & \cellcolor{myg10} 9.36 \\
& ARC-E & 57.62 & 53.63 & 55.82 & 53.95 & \textbf{59.96} & 56.93 & \underline{59.35} & 55.87 & 54.29 & \cellcolor{myg8} 4.09 & \hspace{-3pt}-1.58 \\
& ARC-C & 33.75 & 34.60 & 34.68 & 35.17 & 38.90 & 39.14 & 38.64 & \underline{39.50} & \textbf{43.39} & \hspace{-2.5pt}-0.60 & \cellcolor{myg8} 3.89 \\
\cline{1-2}
\multirow{6}{*}{\makecell[l]{\textbf{Under-}\\\textbf{standing}}} & Winogrande & 68.01 & 64.97 & 67.83 & 65.21 & \textbf{71.38} & 69.82 & \underline{71.03} & 69.05 & 69.28 & \cellcolor{myg7} 2.33 & \cellcolor{myg4} 0.23 \\
& OpenbookQA & 77.10 & 73.28 & 77.15 & 72.12 & \textbf{81.73} & 78.31 & \underline{81.29} & 77.51 & 81.05 & \cellcolor{myg9} 4.22 & \cellcolor{myg7} 3.54 \\
& \mbox{Race-Middle} & 63.99 & 60.10 & 62.84 & 59.45 & \textbf{74.82} & 68.25 & 72.06 & 68.34 & \underline{74.24} & \cellcolor{myg9} 6.48 & \cellcolor{myg9} 5.90 \\
& Race-High & 58.74 & 53.24 & 54.91 & 52.69 & \textbf{71.05} & 59.20 & 65.67 & 57.72 & \underline{69.62} & \cellcolor{myg12} \hspace{-2.5pt}13.33 & \cellcolor{myg11} \hspace{-3pt}11.90 \\
& WSC & 51.30 & 47.21 & 51.06 & 47.72 & \underline{56.17} & 54.18 & 57.30 & 55.23 & \textbf{66.35} & \cellcolor{myg5} 0.94 & \cellcolor{myg11}\hspace{-3pt}11.12 \\
& CHID & 39.05 & 49.66 & 45.26 & 53.49 & 71.94 & 71.82 & 72.92 & \textbf{74.29} & \underline{74.06} & \hspace{-2.5pt}-2.35 & \hspace{-3pt}-0.23 \\
\cline{1-2}
\multirow{4}{*}{\textbf{Reasoning}} & HellaSwag & 63.11 & 63.08 & 62.58 & 63.02 & \textbf{65.70} & 61.90 & 64.32 & 63.24 & \underline{65.12} & \cellcolor{myg7} 2.46 & \cellcolor{myg6} 1.88 \\
& SIQA & 42.37 & 44.06 & 43.27 & 44.52 & 45.57 & \underline{50.20} & 46.83 & \textbf{51.71} & 49.64 & \hspace{-2.5pt}-6.14 & \hspace{-3pt}-2.07 \\
& PIQA & 71.92 & 71.95 & 70.35 & 71.84 & \underline{76.59} & 74.60 & 73.77 & 75.19 & \textbf{78.06} & \cellcolor{myg6} 1.40 & \cellcolor{myg6} 2.87 \\
& OCNLI & 33.89 & 37.74 & 39.41 & 40.46 & 49.73 & 48.31 & 48.07 & \underline{50.29} & \textbf{50.39} & \hspace{-2.5pt}-0.56 & \cellcolor{myg4} 0.10 \\
\cline{1-2}
\textbf{Math} & GSM8K & 25.19 & 23.72 & 22.68 & 23.05 & 56.77 & 50.10 & 54.25 & \underline{51.37} & \textbf{52.08} &\cellcolor{myg9} 5.40 & \cellcolor{myg5} 0.71 \\
\cline{1-2}
\textbf{Code} & MBPP & 13.80 & 11.29 & 13.92 & 10.80 & \underline{37.50} & 34.82 & 36.72 & 33.20 & \textbf{38.92} & \cellcolor{myg8} 4.30 & \cellcolor{myg9} 5.72 \\
\midrule
\multirow{4}{*}{\textbf{Multimodal}} & \mbox{MMMU-VAL} & -- & 35.67 & 30.78 & 35.56 & \cellcolor{white} -- & 34.56 & 32.33 & 35.11 & \textbf{39.89} & -- & \cellcolor{myg8} 4.78 \\
& \mbox{MMMU-TEST} & -- & 34.40 & 30.90 & 35.33 & \cellcolor{white} -- & 34.90 & 31.80 & 35.10 & \textbf{37.30} & -- & \cellcolor{myg6} 2.20 \\
& MMBench & -- & 63.18 & 59.83 & 65.14 & \cellcolor{white} -- & 66.05 & 62.84 & \textbf{70.94} & \underline{70.53} & -- & \hspace{-2.5pt}-0.41 \\
& ScienceQA & -- & 67.72 & 64.49 & 71.50 & \cellcolor{white} -- & 74.26 & 69.09 & \underline{74.89} & \textbf{78.76} & -- &\cellcolor{myg7} 3.87 \\
\bottomrule
\end{tabular}
\caption{\textbf{Performance comparisons of {\scshape Wings} and the baseline MLLMs under the same training data}.
We consider $8$ baseline MLLMs, including LLMs as $\text{Vicuna}_{\text{v1.5}}$ \& Qwen1.5, visual encoders as CLIP~\cite{radford2021clip} \& SigLIP~\cite{zhai2023siglip}, and training strategies as full-parameter \& LoRA fine-tuning.
The first entry represents the initial LLM, upon which each MLLM is trained.
Our evaluation spans $6$ domains with $20$ datasets.
\Wings is based on the Qwen1.5 and SigLIP, and the column ``Our Improvement'' highlights how much \Wings surpasses its baseline with the same backbones.
}
\label{tab:bench_text_only}
\vspace{-10pt}
\end{table*}

\textbf{Model Summaries \& Implementation Details.} We release the $\text{\Wings}_{\texttt{base}}$ and $\text{\Wings}_{\texttt{pro}}$, with Qwen1.5-$7$B LLM~\cite{qwen} and SigLIP~\cite{zhai2023siglip} visual encoder as the foundations. We also introduce the $\text{\Wings}_{\text{1.8B}}$ version, adapted to Qwen1.5-$1.8$B LLM for edge device compatibility.
As illustrated in~\autoref{fig:training}, we only optimize the projector and the image learners of \Wings for the first alignment stage. The LLM branch adaptation is incorporated during the second instruction tuning stage. We train for $1$ epoch with the AdamW optimizer and the Cosine learning schedule. Typically, the learning rates for the first and second stages are set at $1 e^{-3}$ and $2 e^{-6}$ (with the projector part as $1 e^{-5}$), respectively.
For $\text{\Wings}_{\texttt{base}}$, approximately $1$m training data to align image learners and about $0.6$m supervised fine-tuning instructions for the next stage (the same as LLaVA$_{\text{v1.5}}$~\cite{liu2023llava}).
In the $\text{\Wings}_{\texttt{pro}}$, we use the same aligned data and approximately $2$m training data for learning image-text learners.
These two types of MLLM require about $1.5$ and $6$ days of training on $8 \times$ A100 GPUs, respectively.
The training datasets for $\text{\Wings}_{\texttt{mini}}$ are consistent with the $\text{\Wings}_{\texttt{pro}}$. It takes approximately $5$ days to run on $4 \times$ A100 GPUs.

\begin{table*}[t!]
\scriptsize
\vspace{-20pt}
\centering
\setlength\tabcolsep{1pt}
\renewcommand{\arraystretch}{1.34}
\begin{tabular}{ p{7.7em} p{4.9em}<{\centering} p{4.9em}<{\centering} p{2.6em}<{\centering} p{2.5em}<{\centering} p{2.5em}<{\centering} p{2.5em}<{\centering} p{2.5em}<{\centering} | p{4.9em}<{\centering} p{4.9em}<{\centering} p{2.8em}<{\centering} p{2.5em}<{\centering} p{2.5em}<{\centering} p{2.5em}<{\centering} p{2.5em}<{\centering}}
\toprule
\multirow{2}{*}{\diagbox[]{\textbf{Method}}{\textbf{Dataset}}} & \multicolumn{7}{c|}{\textbf{Text-Only QAs}} & \multicolumn{7}{c}{\textbf{Multimodal QAs}} \\
& \multicolumn{7}{l|}{\tiny{\textbf{MMLU/C*}}\,\,\,\,\,\tiny{\textbf{RACE-M/H}}\,\,\,\,\tiny{\textbf{ARC}}\,\tiny{\textbf{HellaSwag}}\,\tiny{\textbf{Winog.}}\,\tiny{\textbf{GSM8K}}\,\tiny{\textbf{MBPP}}} & \multicolumn{7}{l}{\tiny{\textbf{MMMU-V/T\,\,MMB-EN/CN\,\,\,MME\,\,\,\,SciQA\,\,\,\,SEED\,\,\,\,\,AI2D\,\,\,\,HallB}}} \\
\midrule
& & & \multicolumn{12}{l}{\em Equal-Scale Open-Source $7$B Multimodal LLMs}\\
\noalign{\vskip 0.7ex}
\cline{4-9}
\noalign{\vskip 0.7ex}
O-Flamingo$_{\text{v2}}$~\cite{DBLP:journals/corr/abs-2308-01390} & 26.9 / 27.1 & 40.3 / 32.6 & 31.0 & 55.4 & 58.3 & 10.2 & \,\,9.1 & 29.1 / 28.7 & 10.9 / 13.3 & \,\,803.9 & 55.8 & 30.2 & 32.6 & 30.4 \\
IDEFICS~\cite{idefics2023} & 33.0 / 26.4 & 38.2 / 36.9 & 33.2 & 58.9 & 60.2 & 11.7 & \,\,8.1 & 17.6 / 20.2 & 49.6 / 27.3 & 1239.3 & 62.4 & 44.8 & 43.4 & 24.6 \\
InstructBLIP~\cite{instructblip} & 43.2 / 35.7 & 52.8 / 49.7 & 39.5 & 55.7 & 54.9 & 18.3 & 10.3 & 32.7 / 32.1 & 38.5 / 26.8 & 1425.6 & 61.3 & 45.7 & 41.1 & 33.3 \\
ShareGPT4V~\cite{chen2023sharegpt4v} & 47.6 / 36.9 & 55.9 / 51.0 & 41.6 & 54.7 & 60.1 & 18.0 & \,\,8.9 & 35.5 / 35.2 & 67.4 / 63.1 & \textbf{1915.3} & 68.9 & 68.1 & 58.2 & 26.6 \\
Qwen-VL~\cite{bai2023qwenvl} & 49.7 / 58.3 & 65.2 / 64.8 & 34.4 & 58.2 & 61.0 & 49.0 & 34.6 & 36.4 / 35.9 & 60.3 / 57.4 & 1806.2 & 69.6 & 62.0 & 61.9 & 34.1 \\
Monkey~\cite{li2023monkey} & 52.8 / 66.9 & 65.6 / 62.1 & 38.2 & 60.6 & 59.3 & 51.8 & 37.1 & \textbf{40.3} / \underline{37.1} & 71.9 / 67.8 & \underline{1815.4} & 78.3 & 69.1 & 62.5 & 42.1 \\
LLaVA$_{\text{v1.5}}$~\cite{liu2023llava} & 51.1 / 38.3 & 60.1 / 53.2 & 34.6 & 63.1 & 65.0 & 23.7 & 11.3 & 35.7 / 34.4 & 63.2 / 57.7 & 1518.6 & 67.7 & 63.7 & 56.4 & 29.7 \\
LLaVA$_{\text{Next}}$~\cite{liu2024llavanext} & 50.2 / 39.7 & 65.1 / 58.3 & 36.0 & 63.7 & 68.9 & 30.3 & 23.0 & 37.6 / 35.8 & 67.8 / 61.8 & 1760.3 & 70.1 & 69.1 & \underline{66.4} & 29.6 \\
\mbox{DeepSeek-VL}~\cite{lu2024deepseekvl} & 53.9 / 64.0 & 70.6 / 63.8 & 39.2 & \underline{65.1} & 67.2 & \underline{55.3} & \textbf{43.1} & 37.6 / 35.3 & \underline{72.7} / \textbf{72.5} & 1716.8 & \underline{80.6} & \underline{70.0} & \textbf{66.5} & 36.2 \\
\midrule
\rowcolor{myred}
\textbf{{\scshape Wings}} (Ours) & \underline{60.5} / \textbf{69.8} & \underline{74.2} / \underline{69.6} & \underline{43.4} & \underline{65.1} & \underline{69.3} & 52.1 & 38.9 & \underline{39.9} / \textbf{37.3} & 70.5 / 68.3 & 1753.8 & 78.8 & 69.5 & 62.7 & \underline{45.8} \\
\rowcolor{mydred}
\text{\textbf{{\scshape Wings}}}$_{\texttt{\textbf{pro}}}$(Ours) & \textbf{61.3} / \underline{68.5} & \textbf{82.8} / \textbf{76.3} & \textbf{46.3} & \textbf{69.2} & \textbf{70.9} & \textbf{56.3} & \underline{39.3} & 38.2 / 36.9 & \textbf{73.1} / \underline{69.0} & 1786.1 & \textbf{83.1} & \textbf{70.2} & 65.8 & \textbf{47.3} \\
\midrule
& & & \multicolumn{12}{l}{\,\,\,\,\,\,\,\,\,\,\,\,\em Advanced Private Multimodal LLMs}\\
\noalign{\vskip 0.7ex}
\cline{4-9}
\noalign{\vskip 0.7ex}
GPT-4~\cite{openai2023gpt4} & 83.5 / 71.2 & 93.2 / 87.8 & 93.6 & 88.4 & 75.6 & 91.6 & \,\,56.2$^{\dag}$ & -- & -- & -- & -- & -- & -- & -- \\
GPT-4V~\cite{gpt4v} & 79.3 / 69.4 & 93.7 / 89.2 & 92.9 & 84.7 & 76.1 & 88.4 & 72.4 & 58.9 / 56.8 & 77.0 / 74.4 & 2153.6 & 68.4 & 73.7 & 75.5 & 46.5 \\
Gemini$_{\text{pro vision}}$~\cite{reid2024gemini1_5} & 85.9 / 73.7 & 88.9 / 83.2 & 85.0 & 78.8 & 71.5 & 86.4 & 61.5 & 60.6 / 62.2 & 73.6 / 74.3 & 2193.2 & 58.3 & 70.8 & 70.2 & 45.2 \\
\noalign{\vskip 1.3ex} 
\toprule
& & & \multicolumn{12}{l}{\scriptsize{\,\,\,\,{\textit{Efficient Multimodal LLMs with} \text{\Wings}$_{\text{1.8B}}$}}}\\
\noalign{\vskip 0.7ex}
\cline{4-9}
\noalign{\vskip 0.7ex}
\mbox{\tiny{DeepSeek-VL$_{\text{1.3B}}$}}\tiny{~\cite{lu2024deepseekvl}} & 31.7 / 38.2 & 63.6 / 58.4 & 35.8 & \textbf{52.9} & 45.7 & 17.6 & 16.3 & 33.8 / 32.3 & \underline{65.1} / 60.7 & 1483.4 & \underline{65.4} & \underline{63.3} & 50.1 & 25.0 \\
\mbox{\tiny{MiniCPM-V$_{\text{2.4B}}$}}\tiny{~\cite{minicpm}} & \underline{42.4} / \underline{40.9} & \textbf{68.8} / \underline{62.6} & \underline{37.0} & 48.3 & \underline{51.7} & \underline{32.5} & \underline{24.2} & \textbf{37.2} / \textbf{34.4} & \textbf{65.7} / \textbf{64.1} & \textbf{1584.1} & 64.9 & \textbf{64.7} & \underline{54.9} & \textbf{31.8} \\
\midrule
\rowcolor{mypink}
\mbox{\text{\textbf{{\scshape Wings}}}$_{\text{\textbf{1.8B}}}$(Ours)} & \textbf{44.9} / \textbf{50.9} & \underline{68.5} / \textbf{63.2} & \textbf{37.1} & \underline{50.5} & \textbf{53.0} & \textbf{40.6} & \textbf{28.5} & \underline{35.7} / \underline{33.9} & 64.2 / \underline{61.2} & \underline{1527.3} & \textbf{67.5} & 62.8 & \textbf{55.2} & \underline{30.2} \\
\bottomrule
\end{tabular}
\caption{\textbf{Performance comparisons of the equal-scale MLLMs and the efficient multimodal LLMs} on text-only and multimodal datasets. We evaluate the open-source, efficient, and private API MLLMs. We select $18$ representative evaluation datasets. C* represents the CMMLU dataset.}
\label{tab:bench}
\vspace{-15pt}
\end{table*}

\subsection{Toward Comprehensive Text-only and Multimodal Performance}

\textbf{Text-only Comparison in Fair Data and Resource Environments.}
As shown in~\autoref{tab:bench_text_only}, ``Vicuna-v1.5 + CLIP'' corresponds to LLaVA$_{\text{v1.5}}$, and ``Qwen1.5 + SigLIP'' serves as the foundation for \Wings. When comparing LLM itself and the rest of MLLMs, we observe that fine-tuning with multimodal instructions, compared to the ``Qwen LLM'',  there is text-only forgetting in $12$ out of $16$ datasets, with notable decreases of up to $9.70$, $8.91$, and $13.33$ in MMLU, CMMLU, and RACE-High, respectively.
\Wings significantly improve performance on datasets such as MMLU, CMMLU, RACE-High, and WSC, despite the potential for severe text-only forgetting on baselines. Additionally, we find that the forgetting effects of CLIP and SigLIP are similar. In contrast, parameter-efficient fine-tuning methods like LoRA result in less text-only forgetting but underperform on multimodal questions.
Overall, \Wings' visual and textual learners are credibly demonstrated to retain performance on text-only tasks while also performing well on visual-related questions.
In datasets like CHID, OCNLI, and SIQA, MLLMs show improved text-only performance due to increased language diversity (\eg, Chinese context) or semantic similarity in their fine-tuning data.

\textbf{General Evaluation in Text-Only and Multimodal Tasks.} We present the performance of $9$, roughly $8$B open-source MLLMs, $2$ roughly $2$B, and $2$ private API ones evaluated in the general text-only and multimodal tasks.
\autoref{tab:bench} shows that \Wings series can perform better on text-only and multimodal question-answering datasets. It achieves state-of-the-art performance on 13 out of 18 datasets, significantly surpassing LLaVA$_{\text{v1.5}}$ with the same architecture.
We find that \Wings is equally effective for more efficient foundations, as shown in the ``Efficient Multimodal LLMs'' parts. \Wings can still capture key elements and demonstrate good scalability as the parameter increases.
Although $\text{\Wings}_\texttt{base}$ does not receive additional training for the text-only component, it is still able to achieve comparable performance.

\subsection{Interleaved Image-Text (IIT) Benchmark}

To finely evaluate MLLMs, we construct a series of text-only and multimodal mixed multi-turn conversations. We extract instructions from MMLU, CMMLU, OpenbookQA, HellaSwag, MMMU, MMBench, SEED-Bench, and AI2D datasets with similar semantics by chroma~\cite{chroma}. We then polish the connection between some instructions using GPT-3.5 Turbo to make them closer to real-world conversations.
We set up $6$ vision-content configurations, categorized by the multi-turn content as: (\texttt{T}), (\texttt{T}, \texttt{T}), (\texttt{T}, \texttt{T}, \texttt{T}), (\texttt{T}, \texttt{T}, \texttt{V}), (\texttt{T}, \texttt{V}), and (\texttt{V}). For instance, (\texttt{T}, \texttt{T}, \texttt{V}) indicates two consecutive text-only queries followed by a visual question requiring a response.

\subsection{Ablation Studies}

Referencing \autoref{fig:ablation}, we address three questions to comprehensively analyse \Wings:

\begin{itemize}[itemsep=0pt,topsep=0pt,parsep=0pt,leftmargin=10pt]
    \item Can {\scshape Wings} sustain performance with interleaved evaluation? We find that part (a) highlights \Wings surpassing LLaVA$_{\text{v1.5}}$ and the same-backbone as LLaVA$_{\text{v1.5}}$ (Qwen-SigLIP) for each multi-turn setting, especially in text-centric dialogues.
    \item How do \Wings fare with different learning rate settings? Part (b) demonstrates that using a lower learning rate maintains proficiency in text-only tasks but falls short in multimodal questions, while a higher rate boosts multimodal abilities but not text-only. Applying a higher learning rate to the projector and a lower one to the others achieves the optimal.
    \item Are all components of \Wings equally effective? In part (c), we examine that incorporating visual learners alone slightly preserves text-only abilities, likely by minimizing disruption to the LLM, but diminishes performance on multimodal tasks.
\end{itemize}

In the diverse IIT bench, which ranges from text-rich to multimodal contexts, the effectiveness of \Wings is particularly evident. As shown in \autoref{fig:example}, within real-world applications, textual content offers insights for following visual tasks. \Wings excels in handling text-only tasks while improving performance on visual-related instructions.

\begin{figure}[t]
    \begin{center}
    \vspace{-20pt}
    \centerline{\includegraphics[width=1.1\textwidth]{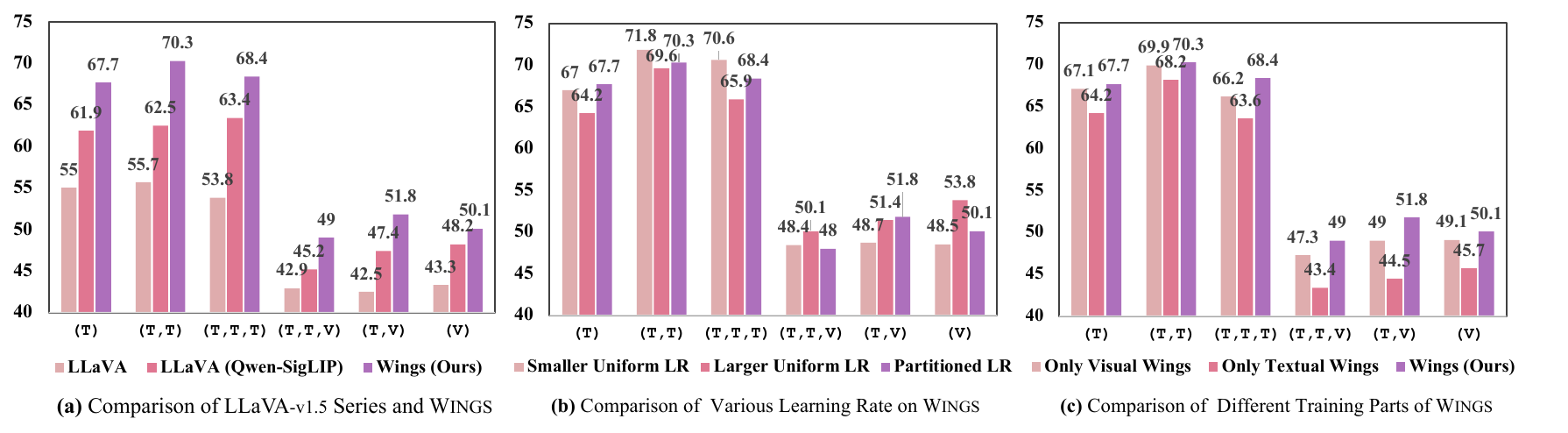}}
    \caption{\textbf{Performance comparison} on the newly constructed \textbf{Interleaved Image and Text (IIT) Benchmark} of the \textbf{LLaVA series}, \textbf{different learning rate} and \textbf{fine-tuning parts}. The horizontal axis represents different multimodal question settings. The horizontal axis shows different multimodal setups, \eg, (\texttt{T}, \texttt{T}, \texttt{I}) represents a visual question after two text-only QAs. The three subfigures represent different ablation settings, with the violet color representing our \Wings.}
    \label{fig:ablation}
    \end{center}
    \vspace{-24pt}
\end{figure}

\section{Conclusion}

We propose \Wings, which includes visual and textual learners, to alleviate text-only forgetting. The learner is composed of efficient Low-Rank Residual Attention (LoRRA).
We start by considering the shifted attention weights in MLLM and, in the first stage, focus on learning the visual learner. Then, we co-train the visual and textual learners with routing based on the shifted attention weights.
\Wings demonstrates remarkable performance on text-only, visual-question-answering, and newly constructed Interleaved Image-Text (IIT) benchmarks. \Wings allows for maintaining text-only performance with limited resources and further enhances performance in well-resourced settings.

\newpage

{\small
\bibliography{main}
\bibliographystyle{plainnat}
}

\appendix
\newpage
\begin{center}
    {\Large{\textbf{Supplementary Material}}}
\end{center}

\section{Experimental Setups and Implementation Details}
\label{supp:training_details}

\textbf{Training Datasets.} The training datasets for the first and second stage of $\text{\Wings}_{\texttt{base}}$ are consistent with LLaVA$_{\text{v1.5}}$~\cite{liu2023llava}. For the second stage, $\text{\Wings}_{\texttt{pro}}$ extends the training dataset to include some visual QA datasets as ALLaVA~\cite{chen2024allava}, SynthDog~\cite{kim2022synthdog}, and ArXivQA~\cite{DBLP:journals/corr/abs-2403-00231}, and text-only QA datasets as Stanford Alpaca~\cite{alpaca}, Alpaca GPT-4~\cite{peng2023instruction}, LIMA~\cite{DBLP:conf/nips/ZhouLX0SMMEYYZG23}, UltraChat~\cite{ding2023enhancing}, WebQA~\cite{DBLP:conf/cvpr/ChangCNGSB22}, and BELLE-0.5M~\cite{belle2023exploring}. $\text{\Wings}_{\text{1.8B}}$ shares the same training set as $\text{\Wings}_{\texttt{pro}}$.

\textbf{Model Structures.} We employ Qwen1.5~\cite{qwen} and SigLIP~\cite{zhai2023siglip} as our foundations.

\textbf{Training Hyperparameters.} We utilize a batch size of $32$, along with the AdamW optimizer and a cosine schedule. For all \Wings-series, the learning rate is set at $1e^{-3}$ for the first stage and adjusts to $2e^{-6}$ for the second stage, except for the projector as $1e^{-5}$.

\textbf{Training Environment.} $\text{\Wings}_{\texttt{base}}$ and $\text{\Wings}_{\texttt{pro}}$ are trained over approximately $1.5$ or $6$ days on $8 \times$ A100 GPUs. $\text{\Wings}_{\text{1.8B}}$ require approximately $5$ days of training on $4 \times$ A100 GPUs.

\section{Additional Experimental Results}

\begin{figure}[http]
    \begin{center}    \centerline{\includegraphics[width=0.55\textwidth]{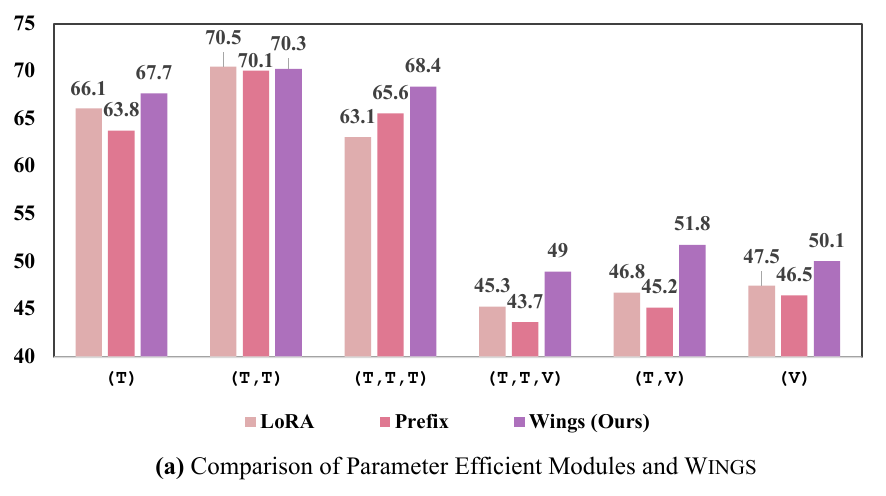}}
    \caption{\textbf{Performance comparison} on the newly constructed \textbf{Interleaved Image and Text (IIT) Benchmark} of the \textbf{Parameter Efficient Modules}. The horizontal axis represents different multimodal question settings. The horizontal axis shows different multimodal setups, \eg, (\texttt{T}, \texttt{T}, \texttt{I}) represents a visual question after two text-only QAs.}
    \label{fig:ablation_supp}
    \end{center}
\end{figure}

Should we only add additional modules on top of an LLM branch or, like \Wings, create two distinct learners for visual and textual modalities? We delve into the low-rank adaptation (LoRA)~\cite{DBLP:conf/iclr/HuSWALWWC22} and Prefix-tuning~\cite{DBLP:conf/acl/LiL20} for minimally adapt to the LLM component.
These techniques introduce optimization parameters beyond the primary branch. These lightweight adjustments align with extensive modifications, effectively minimizing text-only forgetting but concurrently curbing cross-modal positive transfer.

\section{Discussion}

\Wings is a universal plugin that can be integrated with any multimodal mixed-input MLLMs. Notably, it introduces a new concept of competitive reuse among multiple expert groups: we may not require the experts to the Transformer block's $\operatorname{MLP}$ layer at a scale three orders of magnitude larger; instead, a minor update in the attention for better allocation may suffice. This idea is also found in some variants of LoRA~\cite{DBLP:journals/corr/abs-2403-03432,valipour2022dylora}. In the future, we will gradually explore the future of MLLMs.

\section{Limitation \& Broader Impact}
\label{sec:limitation}

Despite \Wings' strong adaptability for embedding auxiliary attention learners in various MLLMs, integrating visual learners requires restarting the feature alignment training, incurring extra costs.
Additionally, its deployment on edge devices faces limitations, with $\text{\Wings}_{\text{1.8B}}$ offering a solution at the expense of performance.
Furthermore, \Wings still requires some text-only data to replay and enhance overall performance, aiming for integration into more generic AI systems in the future.

\begin{figure}[http]
    \begin{center}    \centerline{\includegraphics[width=0.79\textwidth]{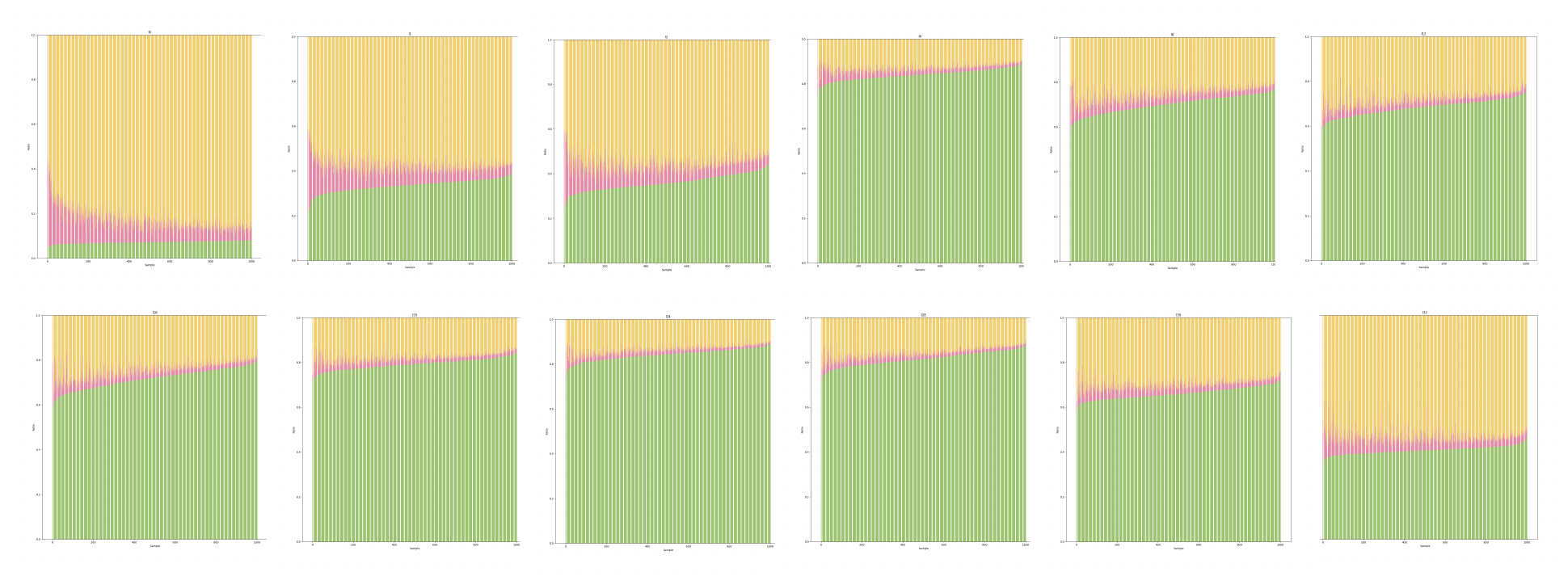}}
    \vspace{10pt}
    \caption{\textbf{Dynamics of Attention Weights from Shallow to Deep Layers.} We calculate the proportion of attention weights for the image-before (yellow), the image-itself (red), and the image-after (green) in each layer. From left to right, top to bottom, from shallow to deep layers.}
    \label{fig:ablation}
    \end{center}
\end{figure}

\begin{figure}[http]
    \begin{center}    \centerline{\includegraphics[width=0.56\textwidth]{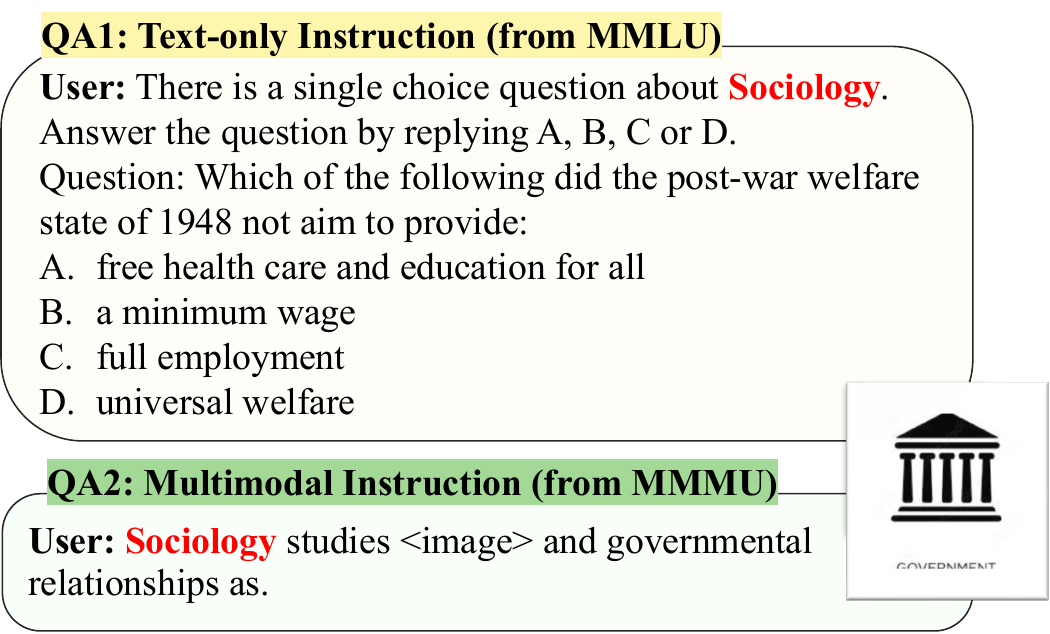}}
    \vspace{10pt}
    \caption{\textbf{An Example of an Interleaved Image-Text Benchmark.} This dialogue is represented as (\texttt{T}, \texttt{V}), consisting of a text-only QA from MMLU~\cite{mmlu} and a visual QA from MMMU~\cite{yue2023mmmu}. It can be observed that, due to the sampling, both include questions from the \textit{Sociology} category.}
    \label{fig:iit_example}
    \end{center}
\end{figure}

\end{document}